\documentclass{article}
\usepackage{framed,multirow}
\usepackage{subfigure}
\usepackage{geometry}
\usepackage{graphicx,amsmath}
\usepackage{url}
\usepackage{hyperref}
\begin{document}
\title{Deep Learning based Person Re-identification}
\author{Nirbhay Kumar Tagore, Ayushman Singh, Sumanth Manche, Pratik Chattopadhyay}
\maketitle
\begin{abstract}
Automated person re-identification in a multi-camera surveillance setup is very important for effective tracking and monitoring crowd movement. In the recent years, few deep learning based re-identification approaches have been developed which are quite accurate but time-intensive, and hence not very suitable for practical purposes. In this paper, we propose an efficient hierarchical re-identification approach in which color histogram based comparison is first employed to find the closest matches in the gallery set, and next deep feature based comparison is carried out using Siamese network. Reduction in search space after the first level of matching helps in achieving a fast response time as well as improving the accuracy of prediction by the Siamese network by eliminating vastly dissimilar elements. A silhouette part-based feature extraction scheme is adopted in each level of hierarchy to preserve the relative locations of the different body structures and make the appearance descriptors more discriminating in nature. The proposed approach has been evaluated on five public data sets and also a new data set captured by our team in our laboratory. Results reveal that it outperforms most state-of-the-art approaches in terms of overall accuracy.
\end{abstract}


\section{Introduction}
Human activity monitoring by means of surveillance cameras is commonly seen in almost all public places, now-a-days. The continuous stream of videos captured by these cameras is manually monitored to detect occurrence of any suspicious events. However, this process of human supervision is time-intensive and error-prone. Over the past two decades, computer vision based researchers have come up with various automated techniques to analyze human video data and extract relevant information for performing human tracking, re-identification, and other vision related tasks.

\begin{figure}[!t]
	\centering
	\includegraphics[scale=.5]{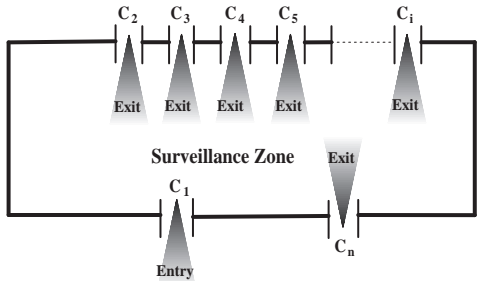}
	\caption{Re-identification scenario}
	\label{fig:scenario}
\end{figure}

In this paper, we focus on short-term re-identification scenario where physical appearance and clothing conditions of a person can be assumed to remain unchanged. Short term re-identification is very useful in tracking subjects while he/she switches position from the field of view of one camera to the other in a multi-camera environment. The re-identification scenario considered in this paper can be explained using Figure \ref{fig:scenario}.
Figure shows a multi-camera surveillance setup in which there are designated entry and exit points. Each of the \emph{n} cameras (labeled as $C_1$, $C_2$, ..., $C_i$, ..., $C_n$) shown in the figure captures walking video of persons from the front view (the gray shaded region associated with each camera represents the field of view of the camera).

We consider a scenario similar to that found in concert and movie halls, or some meeting place. In these places, a group of people assemble in a hall and stay there for a period of time, after which each of them exit the hall one-by-one. For effective tracking of individuals, we propose to install cameras on top of each entry/exit point. A gallery set will formed from all the subjects captured by the several entry gate cameras, which will next be used to re-identify a subject, as he/she approaches an exit gate camera. 
It is easy to comprehend that in the above situation, the physical appearance and clothing conditions of a subject will appear almost same in the videos captured by the two cameras. However, only minor differences in the appearances of subjects might arise due to varying illumination conditions and walking poses. It is also evident that in the above-mentioned scenario, a test subject (i.e., the subject captured by an exit gate camera) will always be a part of the gallery set (i.e., the set of subjects captured by all the entry gate cameras).

Re-identification serves as a convenient way to fuse video information from multiple cameras  (\cite{chattopadhyay2015information})
and perform tracking effectively.

The paper presents a two-level hierarchical 
scheme to compare a test subject against the entire gallery set. 
In the first level, an initial short-listing of the gallery samples is done based on color histogram features 
while in the second level, a pre-trained Siamese network is employed to predict the final class of the test subject from 
the reduced gallery set. 

Our approach has been evaluated on a number of public data sets as well as on a data set captured in the laboratory. 
 Almost in all the different test case scenarios, it has been seen to outperform the state-of-the-art techniques in terms of accuracy and efficiency. 

The important contributions to the paper can be summarized as follows:
\begin{enumerate}
	\item proposing an efficient two-stage hierarchical re-identification algorithm based on color features and deep features obtained from Siamese network
	\item developing a deeper Siamese network architecture by introducing a silhouette part based feature difference layer to improve the accuracy of matching,
    \item carrying out extensive experimental evaluation and comparison with state-of-the-art techniques, and
	\item constructing a re-identification data set of 41 subjects and making it publicly available to the research community.
\end{enumerate}

\section{Related Work}
Automated person re-identification has evolved as an important area of research in computer vision, and 
in general, any re-identification algorithm goes through two major steps: (a) building an appearance model, and (b) comparison using a suitable distance metric (\cite{tu2007intelligent}) (\cite{sivic2006finding}) (\cite{dalal2005histograms}). In the following sub-sections, we present the research trend in person re-identification, 
starting from the non-deep learning based approaches used in the past to the modern deep learning based approaches. 

\subsection{Initial Re-identification Techniques} The re-identification techniques used in the past can be broadly categorized into passive and active approaches. Passive approaches do not involve any feature learning or distance metric learning step (\cite{bazzani2010multiple}, \cite{bazzani2013symmetry}, \cite{forssen2007maximally}). 
These methods usually extract only simple handcrafted features, such as color histograms (\cite{koestinger2012large}, \cite{xiong2014person}), gabor features (\cite{li2013locally}), dense SIFT features (\cite{zhao2013person}), etc., and perform re-identification by comparing these features based on some distance metric. The passive re-identification technique proposed in (\cite{kang2004object}) describes the appearance of a person by means of two different models, both of which are invariant to scaling and other rigid transformations. 
Although the authors of this work have used real surveillance videos in their experiments, they have not presented any thorough quantitative analysis that justifies the effectiveness of their approach. Other passive approaches for re-identification include (\cite{gheissari2006person}), in which feature computation is done from overall appearance of a person, 
(\cite{sivic2006finding}), where a color-based clustering scheme is proposed for short-term re-identification from front view of subjects, and (\cite{corvee2010person}), in which histogram of oriented gradients (HOG) features (\cite{dalal2005histograms}) are employed to detect the human body parts (top, torso, left arm and right arm) for establishing spatial correspondence between individuals.

Distance metric learning based approaches have been used in re-identification to achieve improved performance, such as the large margin nearest neighbor (LMNN-R) classifier (\cite{dikmen2010pedestrian}) that minimizes the distance between true matches and maximizes the distance between false matches. Another similar metric learning approach termed as the relative distance comparison (RDC) is proposed in (\cite{zheng2013reidentification}) in which the authors formulate a logistic function based comparison model for feature quantization to improve the performance of the distance metric learning. In (\cite{ma2012bicov}), the authors proposed extraction of biological inspired features (BIF) from gabor filters, and feature similarity computation using a covariance distance learning approach. 

Active re-identification techniques use either supervised or unsupervised learning strategies for feature descriptor extraction and matching. The brightness transfer function (BTF) introduced in (\cite{javed2005appearance}) establishes appearance correspondences during training. This function computes appearance similarities with the change in color information of a subject while he/she moves from one camera view to another in a non-overlapping camera setup. Another similar brightness transfer function is introduced in (\cite{porikli2003inter}) by Porikli et al. to perform camera color calibration. This is accomplished by employing a dynamic programming based shortest path finding algorithm within the correlation matrix computed from color histograms. 
Active re-identification approaches also make use of descriptor learning, which involves learning of the most discriminative set of features, or applying a weighting scheme to prioritize multiple features. Dimension reduction by discriminative weighting of color, texture, and edge features as described in (\cite{schwartz2009learning}) uses a partial least square (PLS) based approach. 

\subsection{Deep Learning based Approaches}
In the recent years, a number of deep learning based person re-identification techniques have been developed by researchers world-wide. The first among these is the work in (\cite{li2014deepreid}), that comprises of three major components: (a) employing a filter pairing neural network (FPNN) to handle misalignment, occlusion, and background clutter, (b) applying convolutional neural network (CNN) for deep feature extraction, and finally (c) introducing a data set (popularly known as CUHK\_03) based on which the network performance is evaluated. 
A deep framework for scalable distance driven learning is proposed in (\cite{ding2015deep}) which works on relative distance comparison for person re-identification. 
In (\cite{zheng2018subspace}), it is assumed that all images of a person within a particular camera range lie in the same low-rank sub-space, and based on this assumption, a non-negative low-rank and sparse graph is learnt to represent silhouette appearances. Next, NCut is employed to perform sub-space clustering and obtain representative features of a subject in each view. Finally, cross-view quadratic discriminant is used to find the correspondences between the subjects in the two cameras. Another multi-shot based person re-identification technique is described in (\cite{zhou2018easy}) in which reference points based similarity metric is used for pedestrian re-identification.

Siamese neural network has been proposed in (\cite{bromley1994signature}) to evaluate the similarity between two given signature samples. This network provides a similarity score at its output neurons depicting the similarity of the two input signatures, and this network has been successfully applied to research on re-identification recently. For example, the deep metric learning introduced by (\cite{yi2014deep}) for person re-identification use the Siamese network to compare color and texture features. In this work, the authors also perform cross-database experiment to test the robustness of the approach in practical scenarios.  An improvement to the network architecture used in (\cite{yi2014deep}) is proposed in (\cite{ahmed2015improved}), where four convolutional layers are used to obtain deeper features. 
Approaches based on partition-based appearance models for learning include (\cite{li2017learning}) where Siamese Convolutional Neural Network (SCNN) is used to compute features from different body parts, and finally cosine similarity is used to obtain the matching score. 
The architecture proposed in this work is capable of learning both the rigid and latent body part appearances. Additionally, different dilation rates impart a higher degree of robustness to the network, but reduces its efficiency significantly.

An unsupervised approach based on transfer learning of spatio-temporal features is described in (\cite{lv2018unsupervised}) for real-time applications. In this work, a visual classifier is trained on a source data set to learn the spatio-temporal features of pedestrians, in general. These learned features are next fused with spatio-temporal features obtained from the test data using a Bayesian fusion model to perform re-identification. The multi-scale approach described in (\cite{qian2017multi}) deals with differently scaled images of the same person and determines the most suitable scale for matching. The approach in (\cite{lin2017consistent}) aims to maximize the total number of correct matches in a camera network. Here, the authors have used a pre-trained CNN model for feature representation, and next computed a similarity matrix between each pair of subjects using cosine similarity metric. A gradient descent algorithm is followed to maximize the global similarity and minimize certain constraints computed from the similarity matrix, inter-camera inconsistencies, etc. 

Existing deep learning approaches for person re-identification are time-intensive due to feature computation across multiple layers and also due to carrying out comparison with the complete gallery set. 
In this paper, we present a hierarchical person re-identification approach in which the top few best matches are first selected from the gallery set based on appearance features, and next Siamese network based matching is carried out on the reduced gallery set. In contrast to the existing techniques which use Siamese network for image matching, we follow a silhouette part-based matching scheme to preserve the relative locations of the different body parts. Our approach has been seen to perform with higher accuracy than most state-of-the-art techniques. Moreover, due to reduction of the search space after the first level of hierarchy, the average response time of our approach is significantly low.
\section{Proposed Approach} \label{pa}
\begin{figure}[t]
	\centering
	\includegraphics[scale = 0.5]{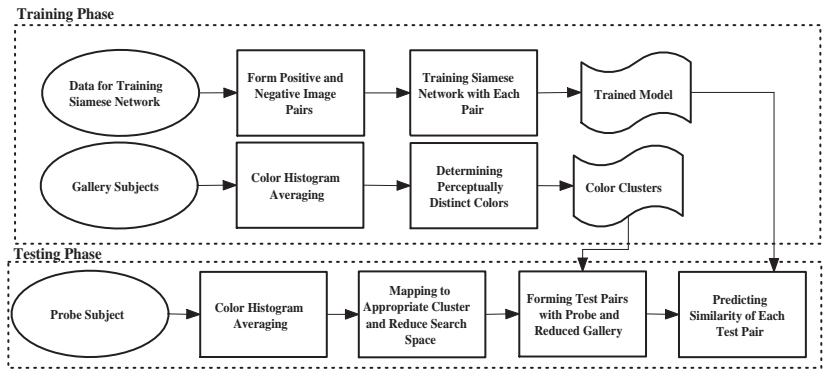}
	\caption{Block diagram of the proposed approach}
	\label{fig:block}
\end{figure}

The proposed re-identification algorithm has been explained taking into consideration two cameras: $C_1$ and $C_2$ positioned on the entry gate and exit gate, respectively (refer to Figure \ref{fig:scenario}).   The algorithm can be conveniently extended to perform re-identification in a multi-camera setup as well. Throughout the text, the walking videos captured by $C_1$ will be considered as the training/gallery set and those captured by $C_2$ will be referred to as the test set. 
 A block diagram of our approach is shown in Figure \ref{fig:block}. 

With reference to this figure, the different training and model-fitting operations involved in the algorithm correspond to the \emph{Training Phase}, while the steps involved in the re-identification of a subject captured by $C_2$ are part of the \emph{Testing phase}. Before training the Siamese network, we use a database of human walking images to construct a large set of image pairs, each of which is labeled as either \emph{positive image pair} or \emph{negative image pair}. Positive image pair implies that both the input images correspond to the same identity, while negative pairs signify that the input images belong to different identities. The trained network is referred to as \emph{Trained Model} in Figure \ref{fig:block}. It may be noted that the database used to train the Siamese network may be different from the gallery set used to perform re-identification. During the deployment phase, 
the set of subjects captured by camera $C_1$ form the gallery set. This gallery set is henceforth used to construct a set of perceptually unique appearance clusters termed as \emph{Color Clusters} based on color histogram feature. As a test subject approaches the camera $C_2$ he is re-identified as one among all the subjects in the gallery set using a two-level hierarchical matching scheme. 
The individual blocks of the re-identification algorithm are explained in detail in the following sub-sections.

\begin{figure}[!t]
	\centering
    \begin{subfigure}[]{
	   \includegraphics[scale=.37]{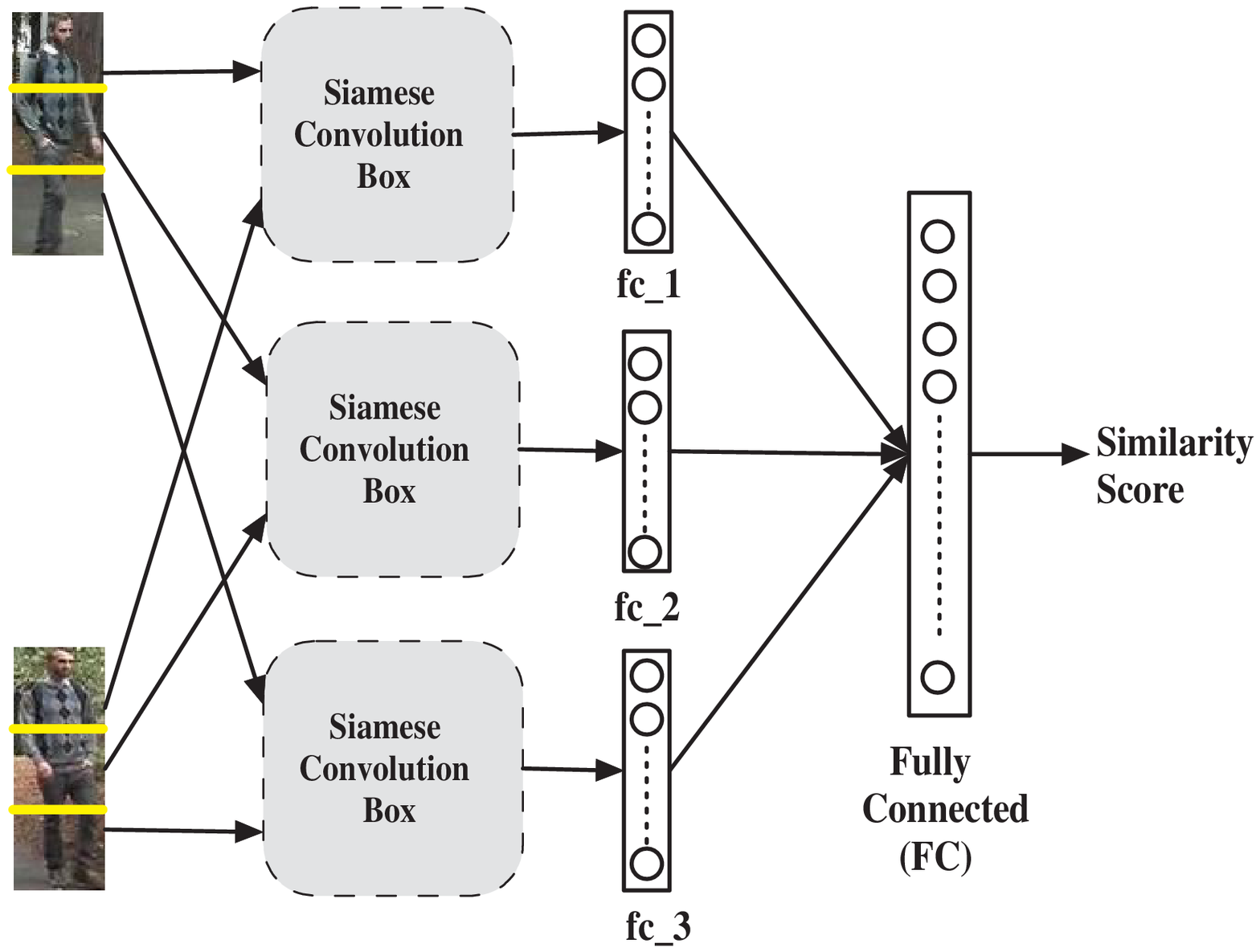}}
    \end{subfigure}
    \begin{subfigure}[]{
	   \includegraphics[scale=.37]{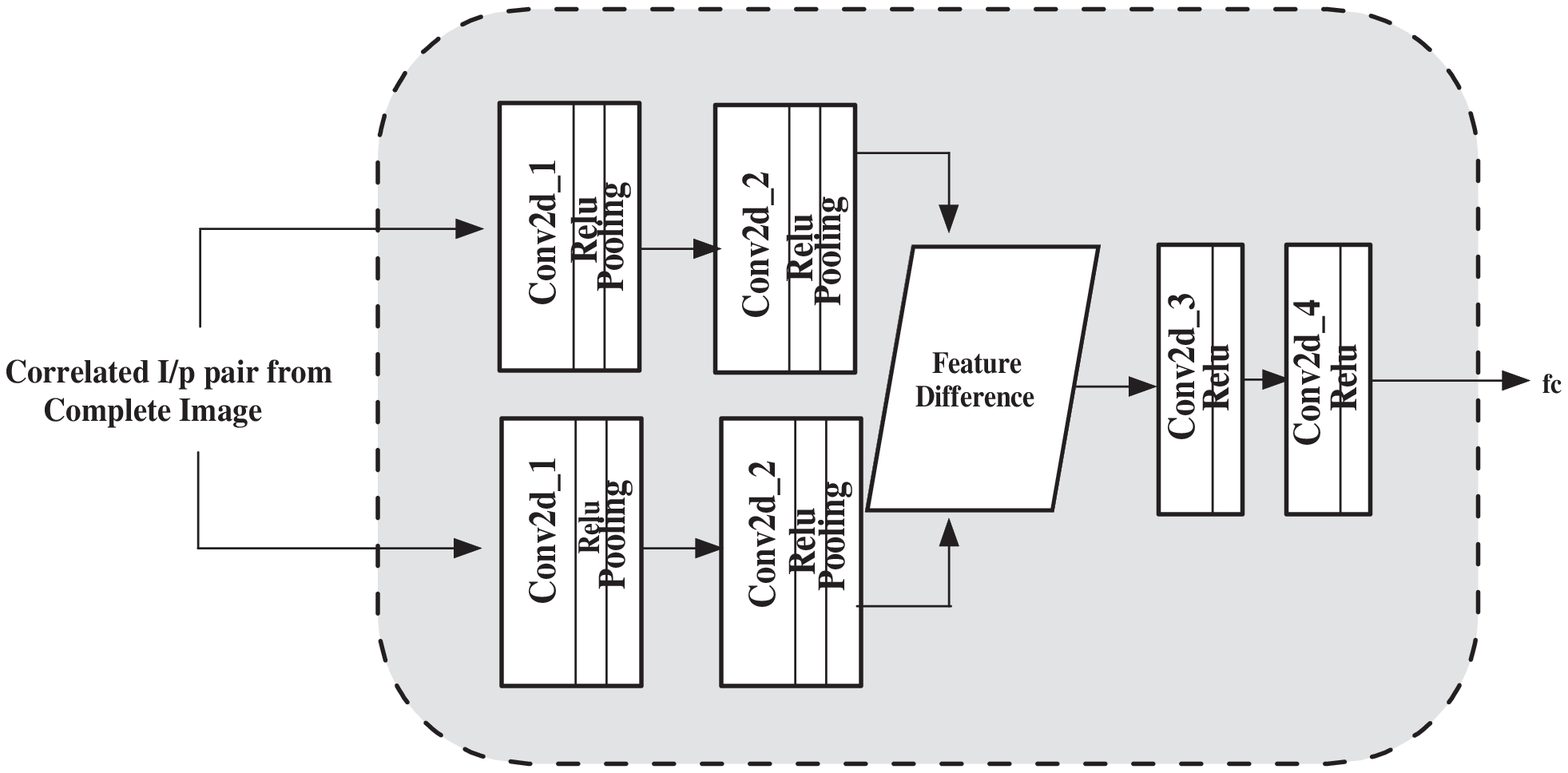}}
    \end{subfigure}
	\caption{(a) Complete architecture of Siamese network (b) Siamese Convolution Box (SCB)}
	\label{fig:SCB}
\end{figure}

\subsection{Training Phase} \label{sct}
\noindent With reference to the block diagram, the first phase of the proposed approach is to train the Siamese network using positive and negative image pairs. The process of constructing the gallery set with positive and negative pairs, as well as training of the Siamese network with this gallery set are discussed next.\\
\noindent\textbf{Forming Positive and Negative Image Pairs}: Suppose the gallery set $S$ captured by camera $C_1$ at the entry gate consists of \emph{N} subjects. 
Also let $N_i$ be the number of frames in the walking sequence of the $i^{th}$ subject, and each frame be denoted by $F^i_j$ (\emph{i} = 1, 2, 3, ..., \emph{N}, and \emph{j} = 1, 2, 3, ..., $N_i$). As discussed earlier, a positive pair is formed from any two frames of same subject, while a negative pair is formed from any two frames of different subjects. Using this process, we generate a large number of positive and negative pairs from a given database using a procedure similar to that described in \cite{ahmed2015improved}. During the Siamese network training phase, the positive pairs are trained with class label \emph{1}, while the negative pairs are trained with class label \emph{0}.\\
\noindent\textbf{Siamese Network Training}:  
Siamese networks are recently being used in the field of person re-identification, since these networks are quite effective in predicting if a pair of input images are similar or not. The Siamese network used in the present study is a four-layer network in which two layers are tied convolved and rest two are normal convolution layers. The complete architecture of the network is shown in Figure \ref{fig:SCB}(a). One of the building blocks of the Siamese network is the Siamese Convolution Box (\emph{SCB}) which is magnified and shown in Figure \ref{fig:SCB}(b). Figure \ref{fig:SCB}(b) shows that after the first two convolution layers, there is a feature difference layer which subtracts the features from the first two layers. This difference feature is further distilled using two more convolution layers without pooling. This helps is preventing significant information loss due to shrinking after the first two layers. The network configuration details are given in Table \ref{Table:net}. As seen in Figure \ref{fig:SCB}(a), three different Siamese blocks are trained on the different silhouette parts, 
namely, the head, torso, and leg part, by dividing the silhouette equally into three regions. The latent vectors obtained from each block (namely, $fc\_1$, $fc\_2$, and $fc\_3$) are compressed to obtain the final fully connected layer (\emph{FC}). The network is trained on an extensive data set constructed from the CUHK\_03 \cite{li2014deepreid} and Market1501 \cite{zheng2015scalable} re-identification data sets. 
As explained before, a label \emph{1} is assigned for positive image pairs, and a label \emph{0} is assigned for negative image pairs. 
The Siamese network is trained in multiple iterations until it achieves a very low error rate $\epsilon$. In our experiments, the value of $\epsilon$ is considered as $10^{-4}$.\\
%
\begin{table}[!t]
	\scriptsize
	\centering
	\caption{Layer specification of SCB network. Both the starting convolution layers are pooled with size 2$\times$2 and the last two layers are exempted from pooling. $\star$ represents the concatenation of fully connected layers}
	\begin{tabular}{|c|c|c|c|}
		
		\hline
		\textbf{Network} & \textbf{Layer} & \textbf{Size of filter} & \textbf{No. of filters} \\ \hline
		\textbf{Siamese} & Conv2d\_1 & 5$\times$5 & 20 \\ 
		\textbf{Convolution}& Conv2d\_2 & 5$\times$5 & 25 \\ 
		\textbf{Box}& Conv2d\_3 & 5$\times$5 & 25 \\ 
		\textbf{(SCB)}& Conv2d\_4 & 3$\times$3 & 25 \\ \hline
		& \textbf{Layer} & \textbf{No. of neurons} &  \\ \cline{2-4}
		\textbf{Fully}& fc\_1 & 500 & - \\ 
		\textbf{Connected}& fc\_2 & 500 & - \\ 
		& fc\_3 & 500 & - \\ 
		& FC & fc\_1$\star$fc\_2$\star$fc\_3 & - \\ \hline
	\end{tabular}
	\label{Table:net}
\end{table}

\noindent\textbf{Color Histogram Averaging}:  As a first step, we aim to group all the gallery subjects into separate clusters based on similar color appearance features. Color histograms in the \emph{R}, \emph{G}, and \emph{B} channels are used as features to perform this grouping. 
For each frame $F^i_j$, we first carry out background subtraction, RGB silhouette extraction, and normalization to fixed height and width using standard techniques \cite{ahmed2015improved},\cite{li2017learning}. 
Since, different silhouettes with similar color distribution are likely to have the same histogram pattern, we first segregate each silhouette into three equal parts as shown in Figure \ref{fig:SCB}(a), and carry out silhouette part-based analysis by computing the histogram for each of the three parts in the \emph{R}, \emph{G}, and \emph{B} channels. Each of these channels is further quantized into 16 bins to eliminate the effect of appearance changes due to illumination differences in the two camera views.  Next, color histograms computed from all the frames of a walking sequence are averaged to generate the final color appearance descriptor $H^i$ of the $i^{th}$ subject. The procedure used to obtain the number of clusters is briefly discussed next.\\
\noindent \textbf{Determining Perceptually Distinct Colors}: The histograms 
computed from all the $N$ subjects (i.e., $H^1$, $H^2$, ..., $H^N$) are concatenated to form a feature matrix $H$, which  
is next clustered into a fixed number of groups using \emph{K}-means clustering. The clustering is done such that subjects with similar perceptual color appearances (i.e., similar $H^î$ features) are placed in the same cluster. The number of color groups (i.e., $K$) to be formed from the set $H$ is determined by plotting an elbow curve. This curve shows the variation of the clustering error (i.e., summation of the square of the intra-cluster distances) as the value of $K$ is gradually increased from a small value. If $\mathcal{C}_1$, $\mathcal{C}_2$, $\mathcal{C}_3$, ..., $\mathcal{C}_K$ are the $K$ cluster centers at a given point of time, then the clustering error is computed as:
\begin{equation}
\centering
    E = \sum_{i=1}^N||H^i - C_{H^i}||^2,
\end{equation}
where  $C_{H^i}$ denotes the cluster center to which $H^i$ gets mapped, and $||.||^2$ denotes the Euclidean norm.
 The elbow curves obtained from the VIPeR and the CUHK\_03 data sets are shown in Figures \ref{optimal}(a) and (b). From each of these figures, it can be seen that the clustering error does not reduce significantly as the number of clusters is increased beyond 100. Hence, a value of $K$ equal to 100 can be considered to be an optimal choice from the elbow curve. Similar approaches to determine the optimal number of clusters can be found in several studies in the past related to diverse applications, e.g., \cite{PEI},\cite{PDV}. 

\subsection{Testing Phase}\label{tep}
\noindent  \textbf{Mapping to Appropriate Color Clusters}:  As soon as a particular subject $S_t^{\prime}$ appears in the field of view of the camera $C_2$ positioned above the exit point, the averaged color histogram (say, $H_t^{\prime}$) of the test subject is computed and the top $\mathcal{K}$ matching clusters are selected. For example, if $\mathcal{C}_1$, $\mathcal{C}_2$, $\mathcal{C}_3$, ..., $\mathcal{C}_\mathcal{K}$ are the top $\mathcal{K}$ matching clusters corresponding to the subject $S_t^{\prime}$, then the subsequent Siamese network based comparison is done on subset $S_{red}$ of $S$ such that $S_{red}$ $\in$ \{$S_{C_1}$$\bigcup$$S_{C_2}$$\bigcup$$S_{C_3}$$\bigcup$... $\bigcup$$S_{C_\mathcal{K}}$\}, where $S_{C_k}$ is the set of subjects belonging to cluster \emph{k}, \emph{k} = 1,2,3,...,$\mathcal{K}$. Reducing the search space by eliminating dissimilar elements helps in improving the prediction accuracy by preventing the subsequent Siamese network based prediction stage from getting biased towards an incorrect element in the gallery set having different color appearance information but closely similar structures.\\
\noindent \textbf{Siamese Network based Prediction}:
Finally, the test subject $S_t^{\prime}$ is compared with all the subjects in the reduced set $S_{red}$ determined from the previous cluster matching stage. Since both $C_1$ and $C_2$ capture videos instead of still images, the average silhouettes computed from the two video sequences are provided as input to the Siamese network. The test subject $S_t^{\prime}$ is assigned the class $\mathcal{T}$ if
\begin{equation}\label{class}
  sim(\mathcal{T},S_t^{\prime}) > sim(X,S_t^{\prime}), \forall X\in{S_{red}}, X\neq\mathcal{T},
\end{equation}
where $sim(A,B)$ represents the similarity score given by the Siamese network for two input images $A$ and $B$.


\section{Data set and Experimental Results}\label{result}

In this section, we present the experiment protocols, data sets used and system configuration details. Five popular public data sets for re-identification as well as a data set captured in our laboratory have been used to evaluate the effectiveness of the proposed work. 
The important characteristics (i.e., number of cameras, total number of images, and number of identities) of each data set are highlighted in Table \ref{tab_data}. 
\begin{table*}[h]
	\renewcommand{\arraystretch}{1.5}
	\centering
	\scriptsize
	\caption{Description of the data sets}\label{tab_data}
	\begin{tabular}{|c|c|c|c|}
		\hline
		\textbf{Dat aset Names} & \textbf{Number of} & \textbf{Number of} & \textbf{Number of} \\
		& \textbf{Cameras} & \textbf{Images} & \textbf{Identities} \\\hline
		\textbf{VIPeR} \cite{gray2007evaluating} & 2 & 1264 & 632 \\ \hline
		\textbf{CUHK\_01} \cite{li2012human}& 2 & 3884 & 971 \\ \hline
		\textbf{CUHK\_03} \cite{li2014deepreid}& 5 pairs & 13160 & 1360 \\ \hline
		\textbf{Market1501} \cite{zheng2015scalable}& 6 & 32268 & 1501 \\ \hline
		\textbf{DukeMTMC-reid} \cite{ristani2016MTMC}& 8 & 36411 & 1812 \\ \hline
		\textbf{IIT(BHU) Re-identification Data Set} & 2 & 1963 &  41 \\ \hline
	\end{tabular}
	\label{Table:data}
\end{table*}

Appropriate citations to each of the data sets (except the last one) used in the study have been stated in Table \ref{Table:data}. It may be noted that, each of these data sets consist of only a few image frames for each subject. To simulate the deployment scenario shown in Figure \ref{fig:scenario}, we have captured a new data set in the laboratory that consists of walking videos of 41 subjects, and made it publicly available to the research community for further comparison. Two different scenarios are considered while preparing the data set: (i) a general situation in which no constraint is imposed on the clothing condition of the subjects, and (ii) a special case situation in which subjects are constrained to wear similar clothes during the video capture phase. The later data set has been captured using a subset of 20 subjects from the previous set of 41 subjects. For each subject, the data set consists of two video sequences captured at different times which simulates the deployment scenario shown in Figure \ref{fig:scenario}. The average number of frames per person in the data set is 48 and the complete size of the uploaded data set is 65.1 MB. This new data set has been termed as the IIT(BHU) Re-identification Data Set (refer to the last row of Table \ref{Table:data}), and the data set can be obtained by clicking on the link; (\href{https://drive.google.com/drive/folders/17JkeKUCBEzUFPYO-Wyaa3VFgBcypuxNG?usp=sharing}{IIT(BHU) Re-identification Data Set}).

\subsection{Experimental Setup}\label{exp}
All experiments have been performed using Tensorflow \cite{abadi2016tensorflow} on a system having 64 GB RAM and NVIDIA TITAN Xp GPU with 12 GB memory capacity. 
Training of the Siamese network for person re-identification (refer to Section \ref{sct}) is done by means of Adam optimizer \cite{kingma2014adam} in multiple epochs until convergence is achieved. Typically,
we terminate the training process when the overall training loss decreases below a pre-defined small threshold (e.g., $10^{-5}$).

In the first experiment, we decide the optimal parameters to train the Siamese network for person re-identification. At each epoch during training, we use the soft-max cross-entropy loss at the final layer of the Siamese network to measure the network error. The only two user-defined constant parameters to be specified before training the Siamese network are the learning rate ($\eta$) and the weight decay factor ($\gamma$). To determine the optimal values for these parameters, we train the network multiple times using \emph{leave-100-out cross-validation} by considering different pairs of $\eta$ and $\gamma$ values, and select the configuration that provides the maximum cross-validation accuracy.

For tuning the network using cross-validation, we consider five different combinations of training and validation sets by randomly splitting a given data set into two parts such that the size of the validation set is always equal to 100. 
The above experiment is repeated for the following three data sets: (a) CUHK\_01 \cite{li2012human}, (b) CUHK\_03  \cite{li2014deepreid} and (c) Market1501 \cite{zheng2015scalable}. In order to get a better estimate of the parameters $\eta$ and $\gamma$, in this experiment we eliminate the search-space reduction step by means of $K$-Means clustering, as discussed in Section \ref{tep}. Each validation sample is compared with the entire gallery set to find the correct match, and the average cross-validation accuracy obtained from all the validation samples is reported in Table \ref{crossval}.

\begin{table}[h]
\renewcommand{\arraystretch}{1.5}
	\centering
	\scriptsize
	\caption{Determination of Siamese network parameters through cross-validation}\label{crossval}
	\begin{tabular}{|c|c|c|c|c|}
		\hline
		\multirow{2}{*}{\textbf{Data set Name}} & \textbf{Number of Training, Test} & \multirow{2}{*}{\textbf{$\eta$}}  & \multirow{2}{*}{\textbf{$\gamma$}} & \textbf{Acc} \\
		& \textbf{and Validation Samples} & &  &\textbf{(\%)} \\
\hline
 		\multirow{6}{*}{\textbf{CUHK\_01} \cite{li2012human}}& 
 		& 0.01 &0.0250 & 62.5 \\
		&No. of Training Samples (771)&0.01&0.0025& 64.8 \\
		&No. of Validation Samples (100)&0.01&0.0005&\textbf{67.4}\\
		&
		&0.04&0.0005& 58.3 \\
		&&0.07&0.0005& 56.7 \\
		 \hline
 		\multirow{6}{*}{\textbf{CUHK\_03} \cite{li2014deepreid}}& 
 		& 0.01 &0.0250& 68.0 \\
 		&No. of Training Samples (1160)&0.01&0.0025& 70.2\\
		&No. of Validation Samples (100)&0.01&0.0005&\textbf{74.6}\\
		&
		&0.04&0.0005& 66.8\\
		&&0.07&0.0005& 62.4\\
 		\hline
 		\multirow{6}{*}{\textbf{Market1501} \cite{zheng2015scalable}}& 
 		& 0.01 &0.0250& 62.7 \\
 		&No. of Training Samples (1301)&0.01&0.0025& 69.2 \\
		&No. of Validation Samples (100)&0.01&0.0005&\textbf{71.8}\\
		&
		&0.04&0.0005& 54.8 \\
		&&0.07&0.0005& 60.4 \\
		 \hline
	\end{tabular}
\end{table}

The first column of the table corresponds to the data set name, the second column shows the number of training samples used for cross-validation, while the third and the fourth columns present the different pairs of $\eta$ and $\gamma$ values considered in the experiments. The final column in the table corresponds to the cross-validated accuracy obtained for each combination. It can be seen from the Table \ref{crossval} that the maximum cross-validation accuracy in case of each of the data sets has been obtained for $\eta$=0.01 and $\gamma$=0.0005. Hence, these values for $\gamma$ and $\eta$ have been considered to train the Siamese network in all the future experiments presented in the paper. Our observation is that the training algorithm converges as the number of epochs reaches 80000 for each of the CUHK\_03 and CUHK\_01 data sets, and 110000 epochs for the Market1501 data. 

\begin{figure}[h]
	\centering
	\begin{subfigure}[]{
			\includegraphics[width=.7\linewidth]{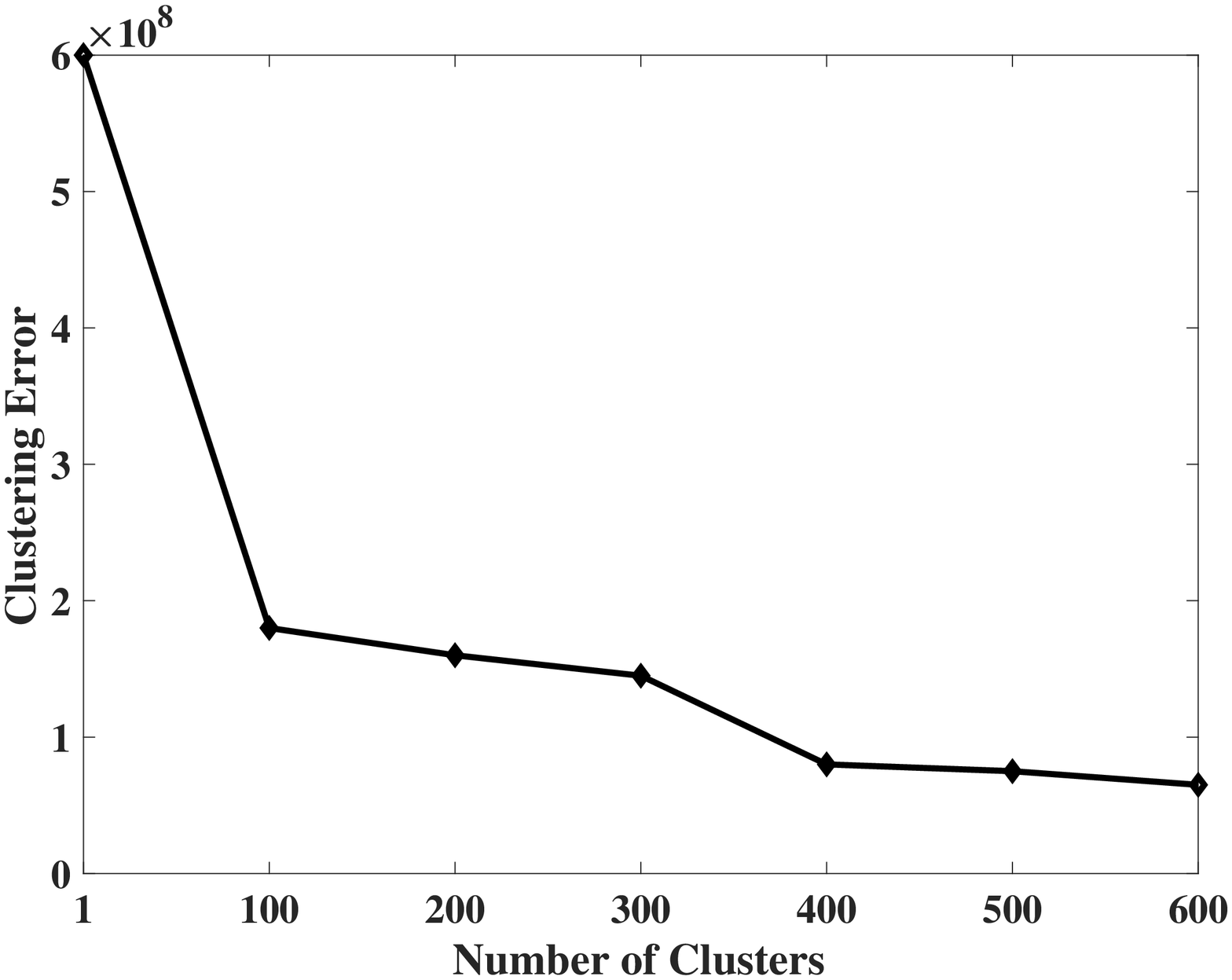}}
	\end{subfigure}
	\begin{subfigure}[]{
			\includegraphics[width=.7\linewidth]{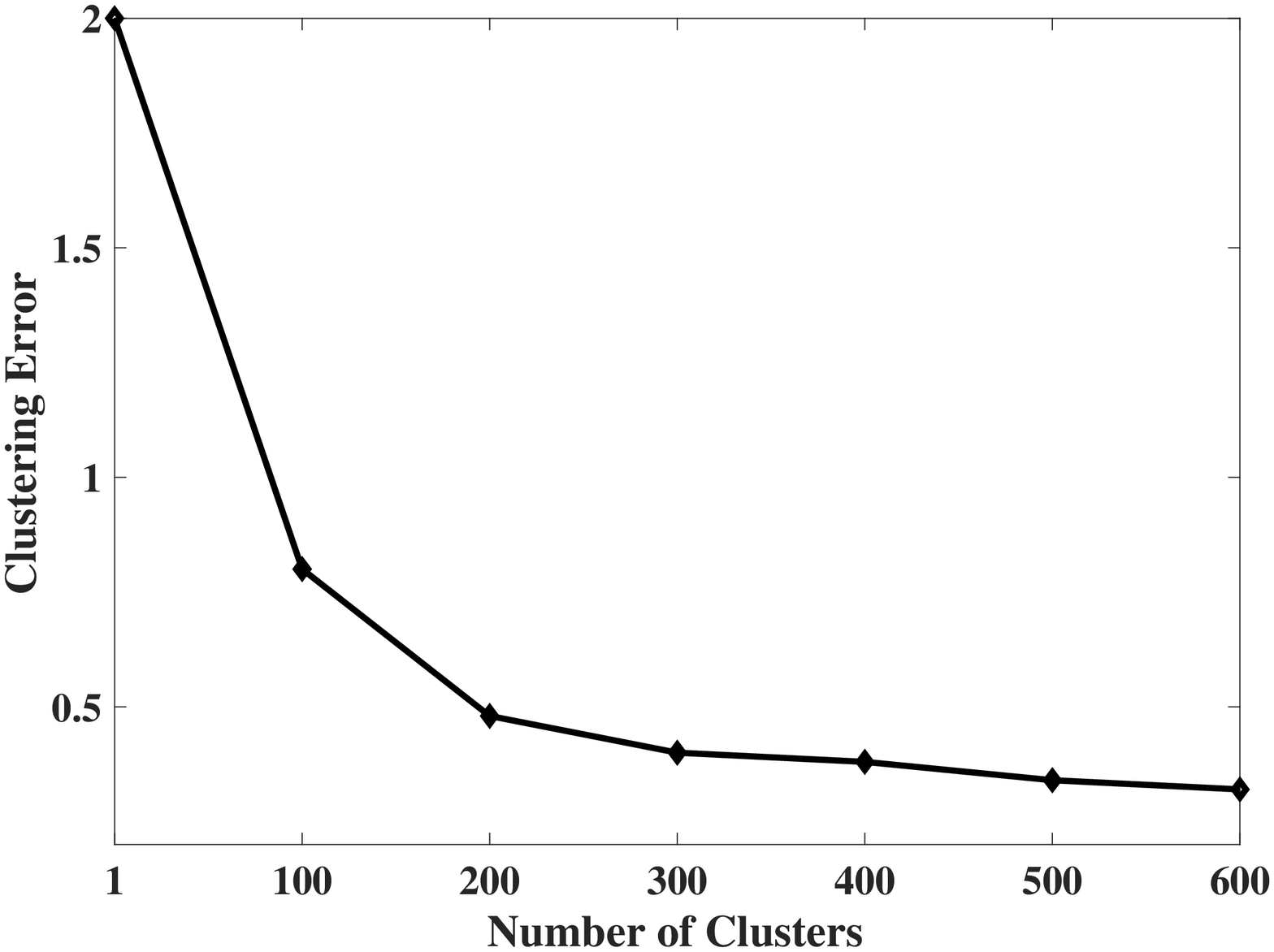}}
	\end{subfigure}
\caption{Elbow Curves for (a) VIPeR Data, (b) CUHK\_03 data}
\label{optimal}
\end{figure}

In our next experiment, we determine the optimal number of perceptually distinct colors from a given data set to perform re-identification. The gallery set for re-identification is formed from one of the data sets shown in Table \ref{Table:data}.
A set of frames corresponding to each subject in the data set is used to prepare this gallery set. The remaining frames form the test set. Improvement in clustering error with increment in the number of clusters (\emph{K}) for the VIPeR and the CUHK\_03 data are shown by means of elbow curves in Figures \ref{optimal}(a) and (b), respectively. 
In each of these figures, 
the horizontal axis represents the number of clusters while the vertical axis denotes the clustering error for the corresponding number of clusters (refer to Section \ref{sct}).  
From both the figures, it is observed that the clustering error is quite low if number of clusters is set to be equal to 100. It is also seen that beyond 200 number of clusters, the curve attains almost a saturation level. Thus, the optimal number of color clusters for the VIPeR data set can be said to lie approximately between 100 and 200. Similar elbow curves drawn for each of the other data sets also reveal that the optimal number of clusters fall in the range [100,200], while in case of the IIT(BHU) Re-identification Data Set, the corresponding number lies within the range [5,10].



\subsection{Deployment Phase} \label{test}
To get a better estimate of the parameters \emph{K} and $\mathcal{K}$ (refer to Sections \ref{sct} and \ref{tep}), in the next experiment we aim to study the effect of varying these parameters on the overall re-identification accuracy.

Table \ref{Table:acc} shows the re-identification accuracy corresponding to a randomly selected sample of 100 test\_ids from every data set given in Table \ref{Table:data}. Results are computed by considering one image sequence from each of these 100 random test ids, and these video sequences were not part of the re-identification gallery set. Accuracy values are reported for the proposed algorithm as the value of \emph{K} is increased from 100 to 500 in steps of 100, and for two values of the parameter $\mathcal{K}$, i.e., 1 and 2. The average time taken to recognize a subject for the different number of clusters corresponding to each data set is also reported in the same table. 
\begin{table}[ht]
	\renewcommand{\arraystretch}{1.5}
	\centering
	\scriptsize
	\caption{Evaluation of the proposed algorithm on public data sets in terms of accuracy (100 test\_ids) and average response time for re-identifying each subject (in milli-secs)} 
	\begin{tabular}{|p{.8cm}|p{.8cm}|p{.3cm}p{.3cm}p{.3cm}p{.35cm}|p{.3cm}p{.3cm}p{.3cm}p{.35cm}|}	
		\hline
		& \textbf{Perf.} &\multicolumn{8}{c|}{\textbf{Selecting Clusters based on Color Matching}}\\
		\cline{3-10}
		\textbf{Data set} &\textbf{Eval.}&\multicolumn{4}{c|}{\textbf{$\mathcal{K}$=1}}&\multicolumn{4}{c|}{\textbf{$\mathcal{K}$=2}}\\
		\cline{2-10}
		& \textbf{Clusters}& \textbf{100} & \textbf{200} & \textbf{300} & \textbf{500} &\textbf{100} & \textbf{200} & \textbf{300} & \textbf{500}\\ \hline
		\textbf{VIPER} & \textbf{Acc}(\%) & 88.3 & 85.4 & 78.6 & 76.5 & 88.0 & 86.4 & 77.5 & 73.8 \\ 
		& \textbf{Time}(\emph{ms}) & 40.5 & 31.2 & 25.5 & 13.6 & 43.8 & 32.4 & 27.6 & 14.1\\ \hline
		\textbf{CUHK01} & \textbf{Acc}(\%) & 82.1 & 83.4 & 76.8 & 72.2 & 84.2 & 85.1 & 78.6 & 75.2\\ 
		& \textbf{Time}(\emph{ms}) & 43.0 & 26.3 & 14.2 & 9.5 & 45.2 & 32.6 & 18.5 & 12.2 \\ \hline
		\textbf{CUHK03} & \textbf{Acc}(\%) & 91.0 & 89.6 & 83.1 & 81.6  & 92.7 & 90.0 & 82.6 & 81.1 \\ 
		& \textbf{Time}(\emph{ms}) & 41.2 & 29.1 & 15.2 & 10.3 & 46.5 & 32.4 & 19.9 & 11.6\\ \hline
		\textbf{Market} & \textbf{Acc}(\%) & 93.6 & 89.4 & 84.8 & 81.0 & 93.2 & 90.7 & 85.6 & 80.3\\ 
		\textbf{1501}& \textbf{Time}(\emph{ms}) & 49.9 & 37.8 & 25.0 & 14.3 & 52.7 & 40.0 & 25.6 & 16.8 \\ \hline
		{\textbf{DukeMT}} & \textbf{Acc}(\%) & 94.0 & 96.3 & 89.2 & 87.6 & 93.8 & 95.0 & 91.2 & 88.7\\ 
		\textbf{MCreid}& \textbf{Time}(\emph{ms}) & 58.8 & 47.7 & 36.6 & 23.4 & 69.1 & 53.4 & 38.1 & 25.5 \\ \hline
	\end{tabular}
	\label{Table:acc}	
\end{table}

It is verified from the table that, as expected, high re-identification accuracy values are always obtained if 
\emph{K} 
is set to be equal to 100 or 200. In general, the accuracy decreases as \emph{K} 
is increased. This is due to the fact that, for higher number of partitions, there is a higher chance for a subject to get mapped to the incorrect color cluster. As explained in Section \ref{tep}, $\mathcal{K}$ is a parameter that decides how many top matching clusters should be retained for the second level of classification using Siamese network. As expected, the response time increases if a higher value of $\mathcal{K}$ is chosen keeping other factors constant. 

Table \ref{Table:acc} shows that, $\mathcal{K}$=2 performs marginally better than $\mathcal{K}$=1. The slightly lower accuracy of $\mathcal{K}$=1 arises due to the fact that sometimes two or more clusters represent similar type of color features and during the cluster mapping stage a test subject gets mapped to an incorrect cluster whose feature vectors are closely similar to that of the correct cluster. Although setting the value of $\mathcal{K}$ to 2 provides marginally better results, it reduces the efficiency of the re-identification process. 
On the other hand, by setting the value of $\mathcal{K}$ to 1, more than 82\% accuracy is obtained for each of the five data sets corresponding to 100 clusters, within a short time. Hence, \emph{K}=100 and $\mathcal{K}$=1 can be treated as optimal parameters for re-identification in case of each of the above data sets. It may be noted that the optimal values for the parameters \emph{K} and $\mathcal{K}$ are data-specific. For a different data set, another set of parameter values might turn out to be optimal. Hence, given any re-identification data set, the optimal values for \emph{K}, and $\mathcal{K}$ has to be first determined before carrying out the re-identification process.

To evaluate the effectiveness of the proposed approach with respect to other state-of-the-art techniques, we next perform a comparative performance analysis of our work with four other recent approaches 
that carry out Siamese network based prediction, namely, 
 \cite{varior2016siamese},\cite{guo2018efficient},\cite{ahmed2015improved},\cite{subramaniam2016deep},
and also two non-Siamese network based techniques: 
Deep-Reid \cite{li2014deepreid}, and MuDeep \cite{qian2017multi}. 
Table \ref{Table:data1} presents the Rank 1 accuracy of the different re-identification approaches on the various data sets used in the study as well as the average time to compare between the two subjects in case of each approach. The first column of the table provides the citations for the different re-identification algorithms used in the study, whereas the second, third and fourth columns present the Rank 1 accuracy obtained by executing each method on the following data sets: (a) IIT(BHU) Re-identification Data Set, (b) VIPeR, (c) CUHK\_01, (d) CUHK\_03. The fifth and sixth columns respectively provide the average accuracy on the different data sets and the average time for comparing between a gallery and a test subject using the corresponding approach. The same training-test pair and the optimal values for the parameters \emph{K} and $K$ determined from the previous experiment have also been used to report the results of this experiment. 
\begin{table}[!h]
	\renewcommand{\arraystretch}{1.5}
	\centering
	\scriptsize
	\caption{Comparison of Rank 1 Accuracy (in \%) for 100 test\_ids of our proposed approach with state-of-the-art techniques along with the average time to compare between two subjects corresponding to each approach}
	\begin{tabular}{|c|c|p{.6cm}|p{.8cm}|p{.8cm}|p{.8cm}|c|}
		\hline
		\textbf{Methods} & \textbf{Our} & \textbf{VIPeR} & \textbf{CUHK01} & \textbf{CUHK03} &\textbf{Avg.}&
		\textbf{Avg. Res.} \\
		&\textbf{Data}&&&&\textbf{Acc(\%)}&\textbf{Time (secs)}\\\hline
		\cite{li2014deepreid}& 67.4 & 56.1 & 27.9 & 26.1 & 44.4 &0.13\\
		\cite{ahmed2015improved} & 72.4 & 35.2 & 64.2 & 55.0& 56.7 & 0.07\\
		\cite{subramaniam2016deep} & 84.6 & 68.7 & 81.2 & 72.3& 76.7 & 0.08\\
        \cite{qian2017multi} & 78.2 & 44.7 & 79.6 & 82.4 & 71.2 &0.06\\
        \cite{guo2018efficient} & 84.7  & 50.9  & \textbf{88.1} & 88.3 & 78.0 &0.08\\
        \textbf{Ours} & \textbf{92.3} & \textbf{91.5} & 87.3 & \textbf{89.6}& \textbf{90.2} & \textbf{0.04}
        \\ \hline
	\end{tabular}
	\label{Table:data1}
\end{table}

\begin{figure*}[t]
	\centering
	\begin{subfigure}[]{
			\includegraphics[width=.3\linewidth]{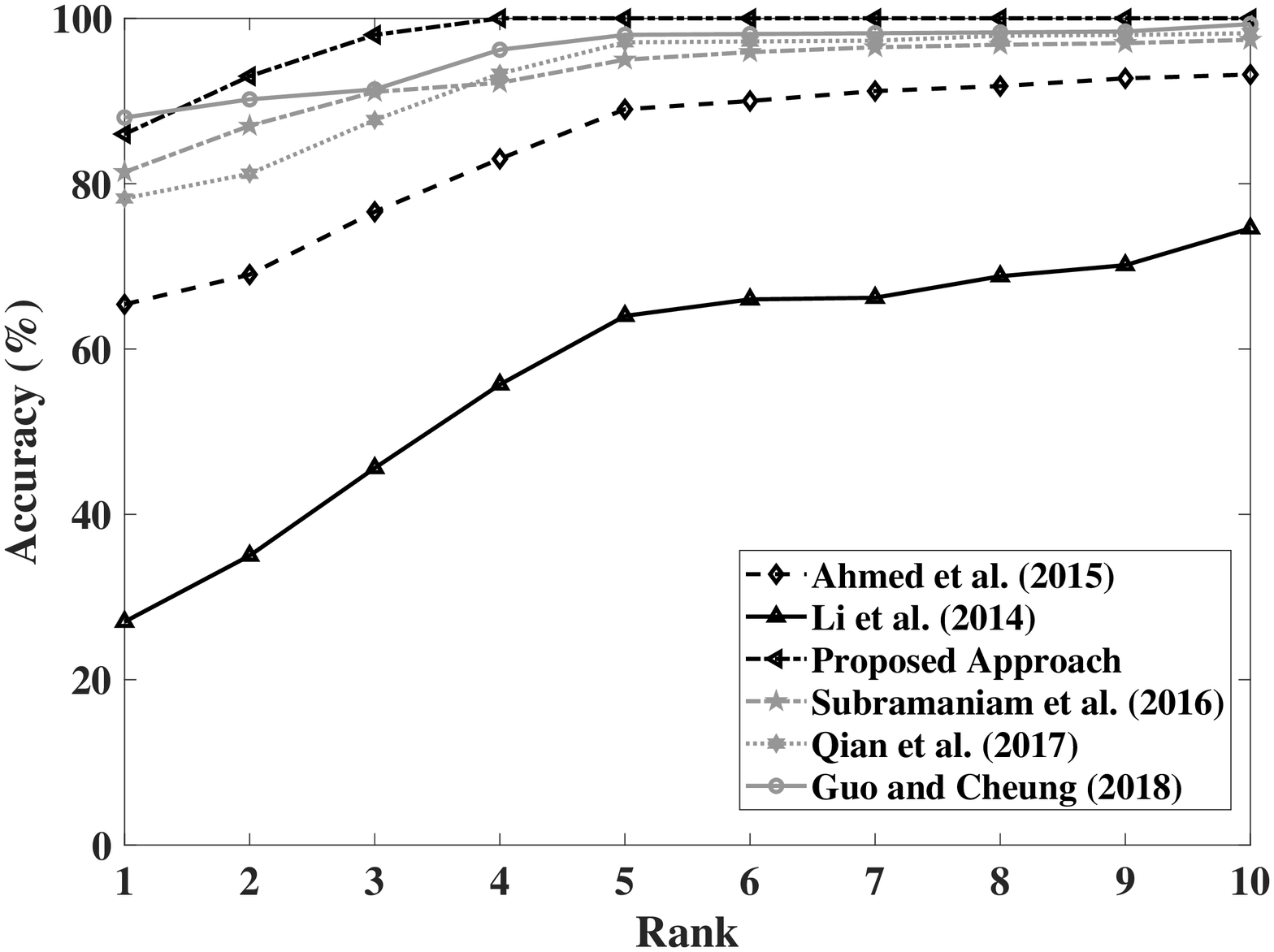}}
	\end{subfigure}
	\begin{subfigure}[]{
			\includegraphics[width=.3\linewidth]{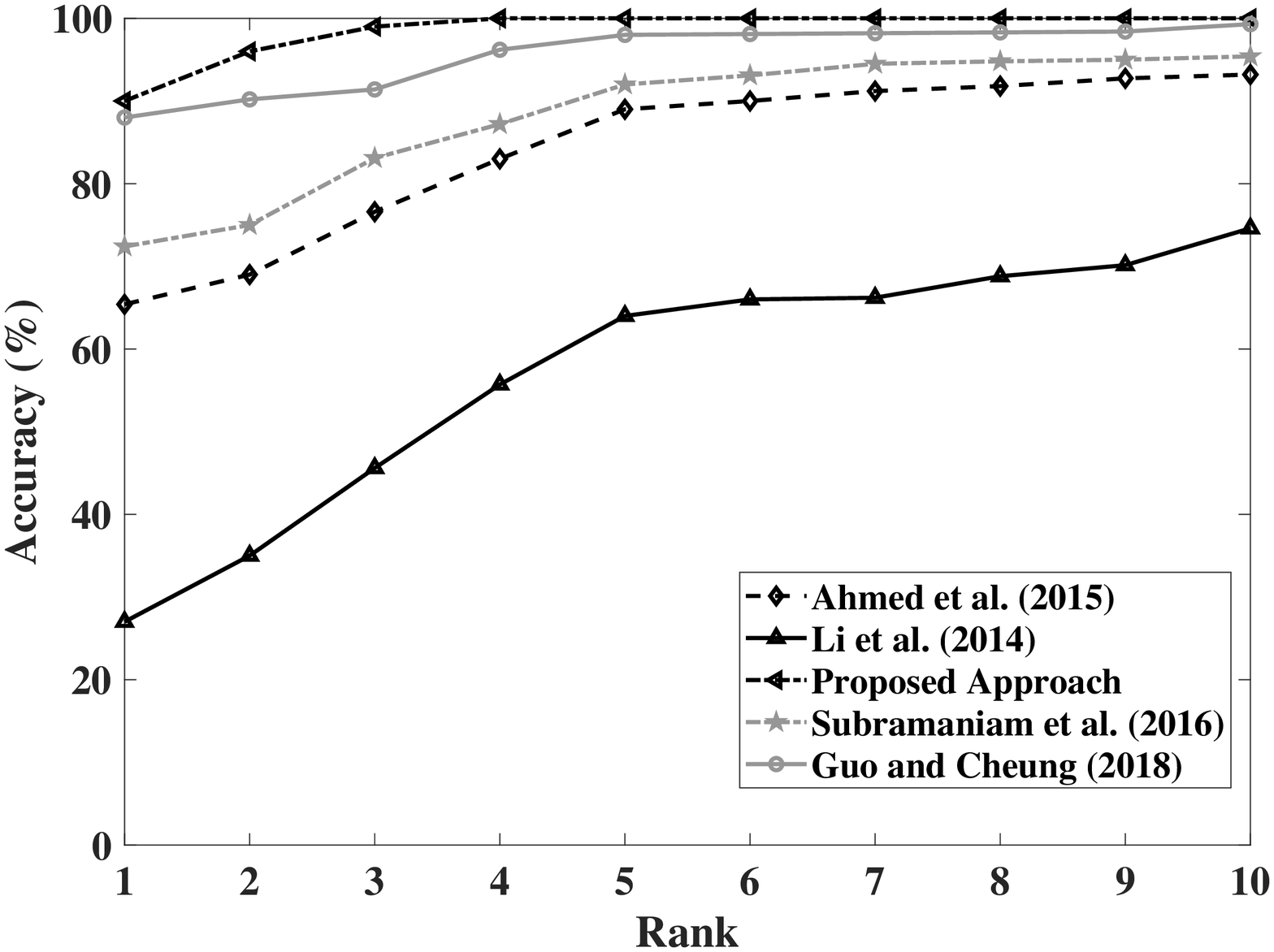}}
	\end{subfigure}
			\begin{subfigure}[]{
			\includegraphics[width=.3\linewidth]{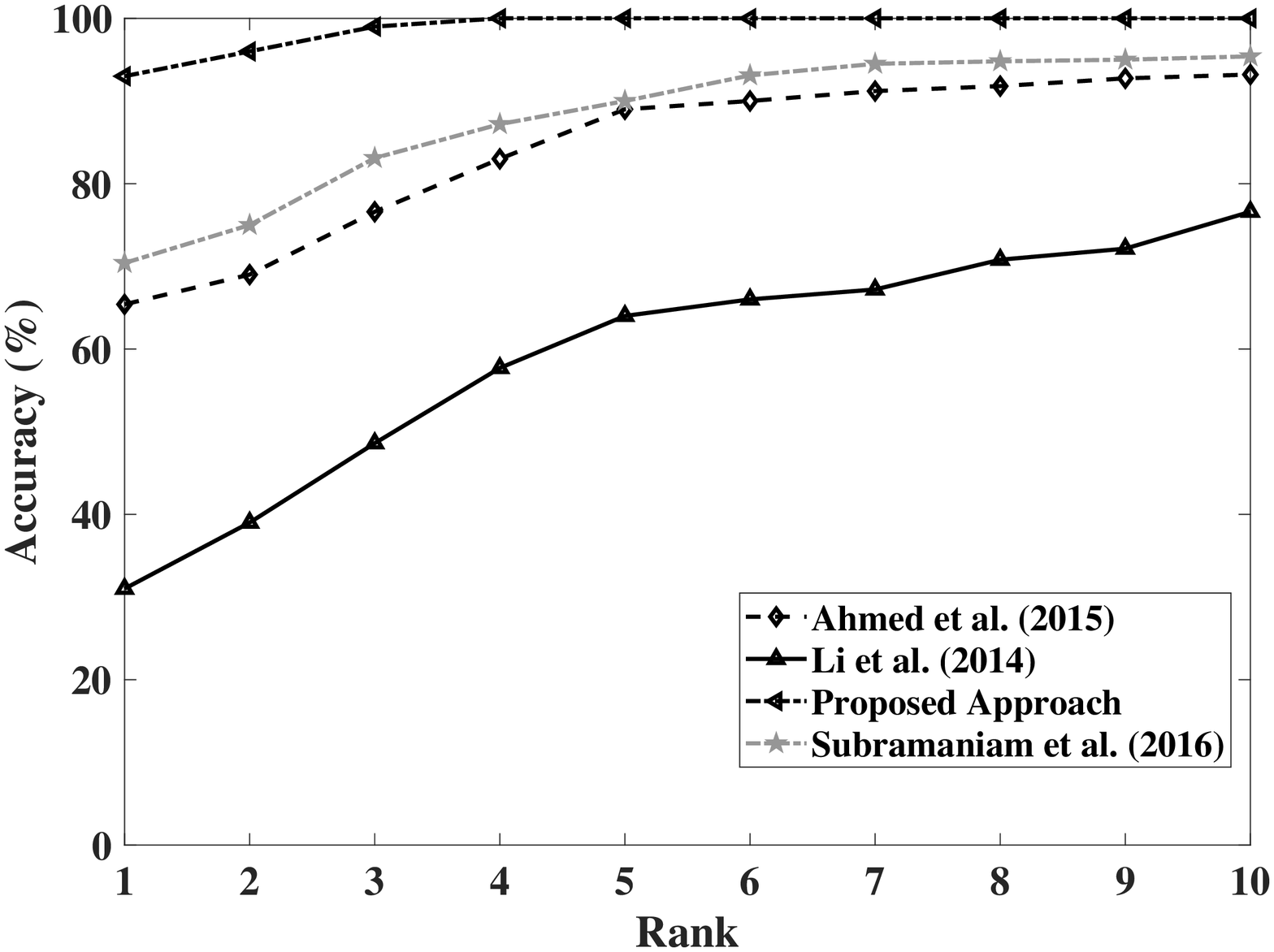}}
	\end{subfigure}
	\begin{subfigure}[]{
			\includegraphics[width=.3\linewidth]{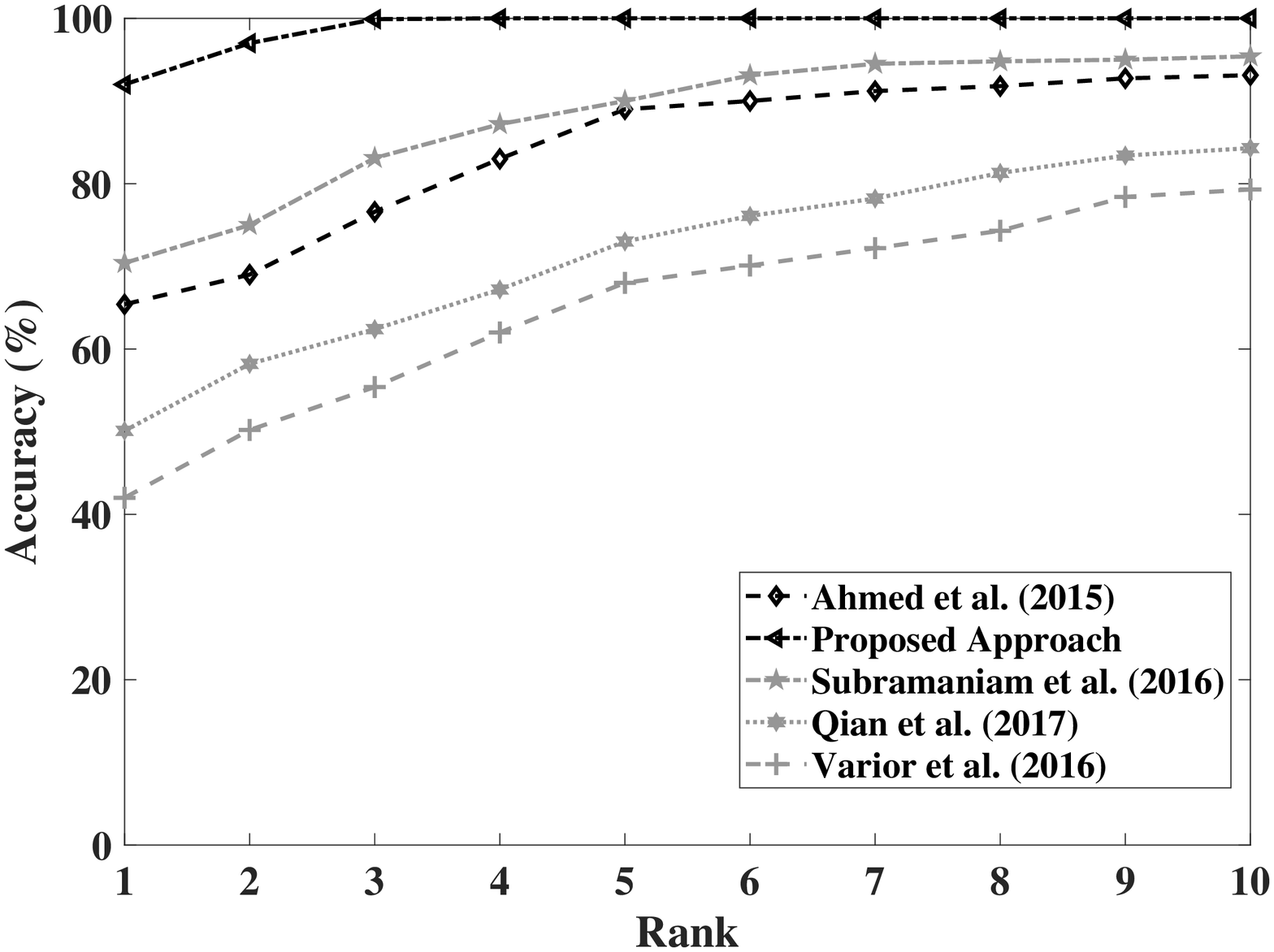}}
	\end{subfigure}
	\begin{subfigure}[]{
			\includegraphics[width=.3\linewidth]{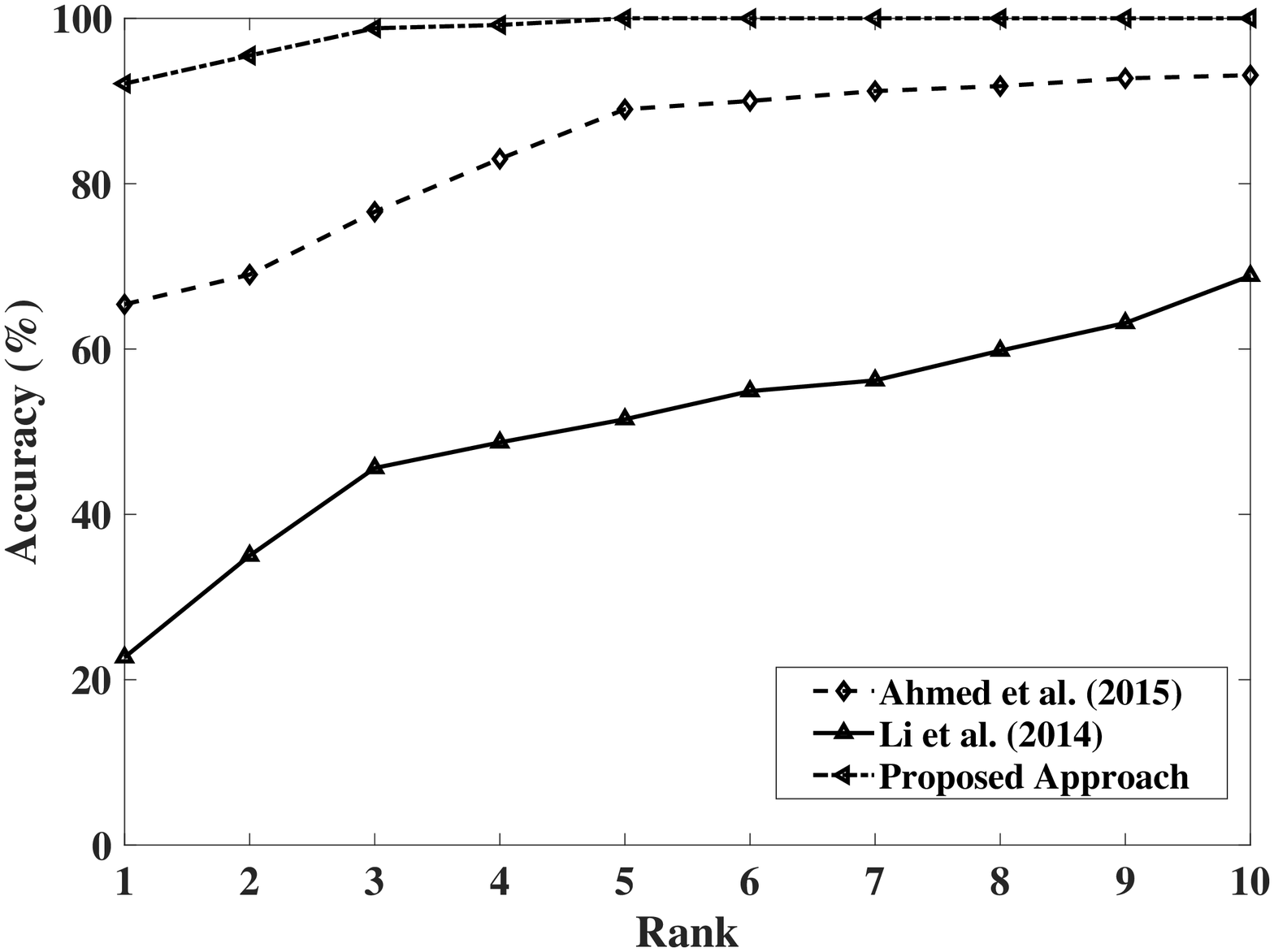}}	
	\end{subfigure}
	\begin{subfigure}[]{
			\includegraphics[width=.3\linewidth]{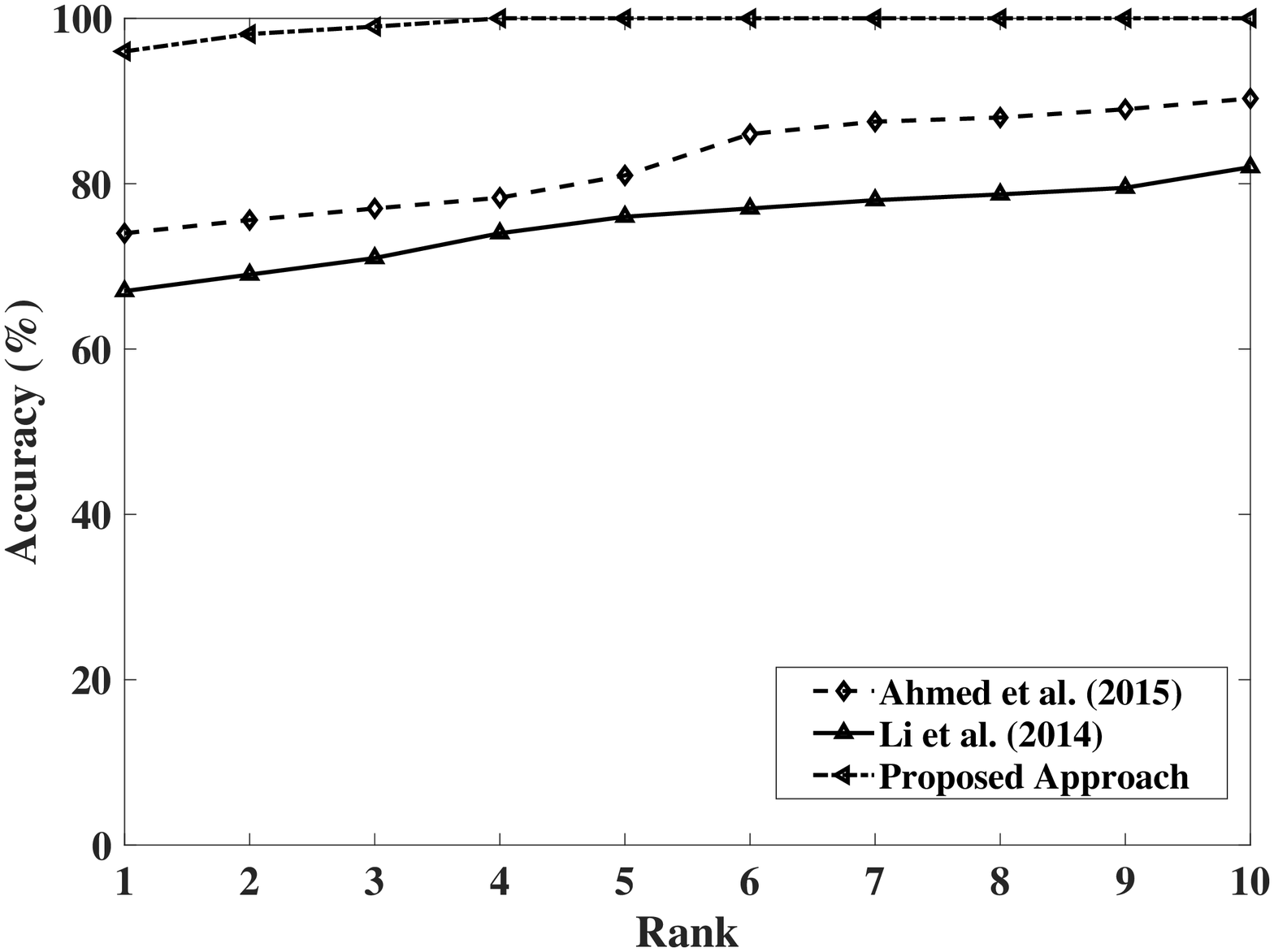}}
	\end{subfigure}
	\caption{CMC on (a) CUHK\_01 Data, (b) CUHK\_03 Data, (c) Market1501 Data, (d) VIPeR Data, and (e) DukeMTMC-reid Data, (f) IIT(BHU) Re-identification Data} 
	\label{cmc}
\end{figure*}

From the table it can be observed that, in general, our approach performs better than the state-of-the-art approaches both in terms of re-identification accuracy and response time. Only in case of the data set CUHK\_01, the Rank 1 accuracy of the approach in \cite{guo2018efficient} shows a slightly higher accuracy. However, the average performance of our approach on the different data sets is considerably better (more than 12\%) than any other existing methods. The superior performance is mostly due to the hierarchical matching scheme followed in this work that eliminates vastly dissimilar candidates after the first stage, thereby preventing the second stage of Siamese network based classification from getting biased towards a different element in the search space. Moreover, the average time to compare between two subjects by applying our approach is only 0.04 seconds, which is more efficient as compared to the state-of-the-art techniques. 

Since Rank 1 accuracy is always not the best measure to evaluate classification performance, we next perform a comparative rank-based performance analysis of our method with other state-of-the-art re-identification approaches by means of 
Cumulative Matching Curves (CMC). 
In each curve, the horizontal axis depicts the rank value from 1 to 10, whereas 
the vertical axis shows the accuracy (in percentage) at a particular rank. Figures \ref{cmc}(a)-(f) respectively present the CMC curves obtained by applying the different re-identification approaches (namely, \cite{ahmed2015improved},\cite{li2014deepreid},\cite{qian2017multi},\cite{guo2018efficient},\cite{varior2016siamese},\cite{subramaniam2016deep}) on the six data sets (refer to Table \ref{Table:data}). 
The rank-based re-identification accuracy obtained for the data sets CUHK\_01 \cite{li2012human}, CUHK\_03  \cite{li2014deepreid} and Market1501 \cite{zheng2015scalable} are presented in Figures \ref{cmc} (a), (b) and (c), respectively. We also perform cross data set experiments and show the rank-based improvement in re-identification accuracy of the different techniques on VIPeR \cite{gray2007evaluating}, DukeMTMC-reid \cite{ristani2016MTMC} and IIT(BHU) Re-identification Data Set in Figures \ref{cmc} (d), (e), (f), respectively.

\begin{figure*}[t]
	\centering
	\begin{subfigure}[ ]{
			\includegraphics[width=.30\linewidth]{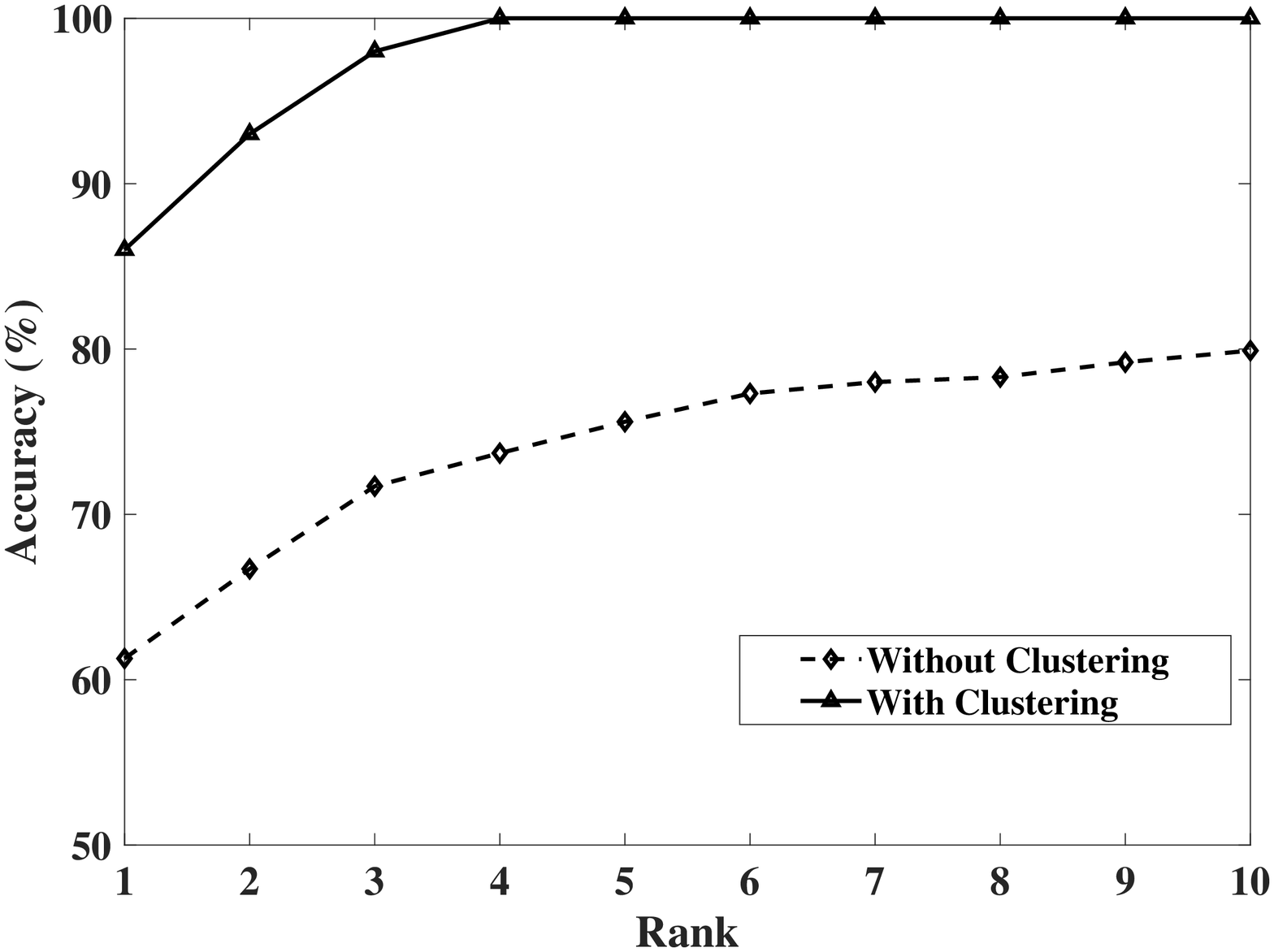}}
	\end{subfigure}
	\begin{subfigure}[ ]{
			\includegraphics[width=.30\linewidth]{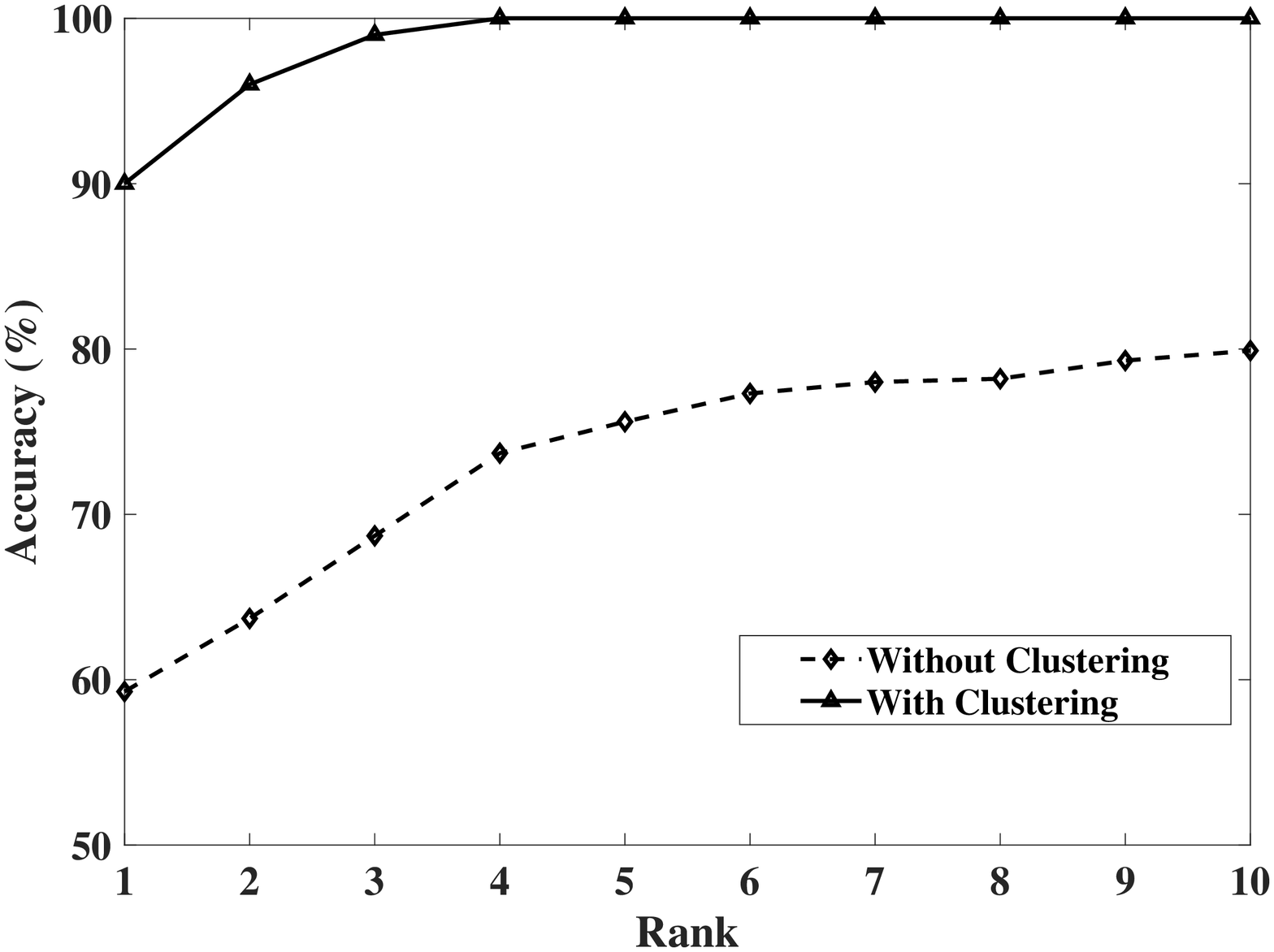}}
	\end{subfigure}
	\begin{subfigure}[ ]{
			\includegraphics[width=.30\linewidth]{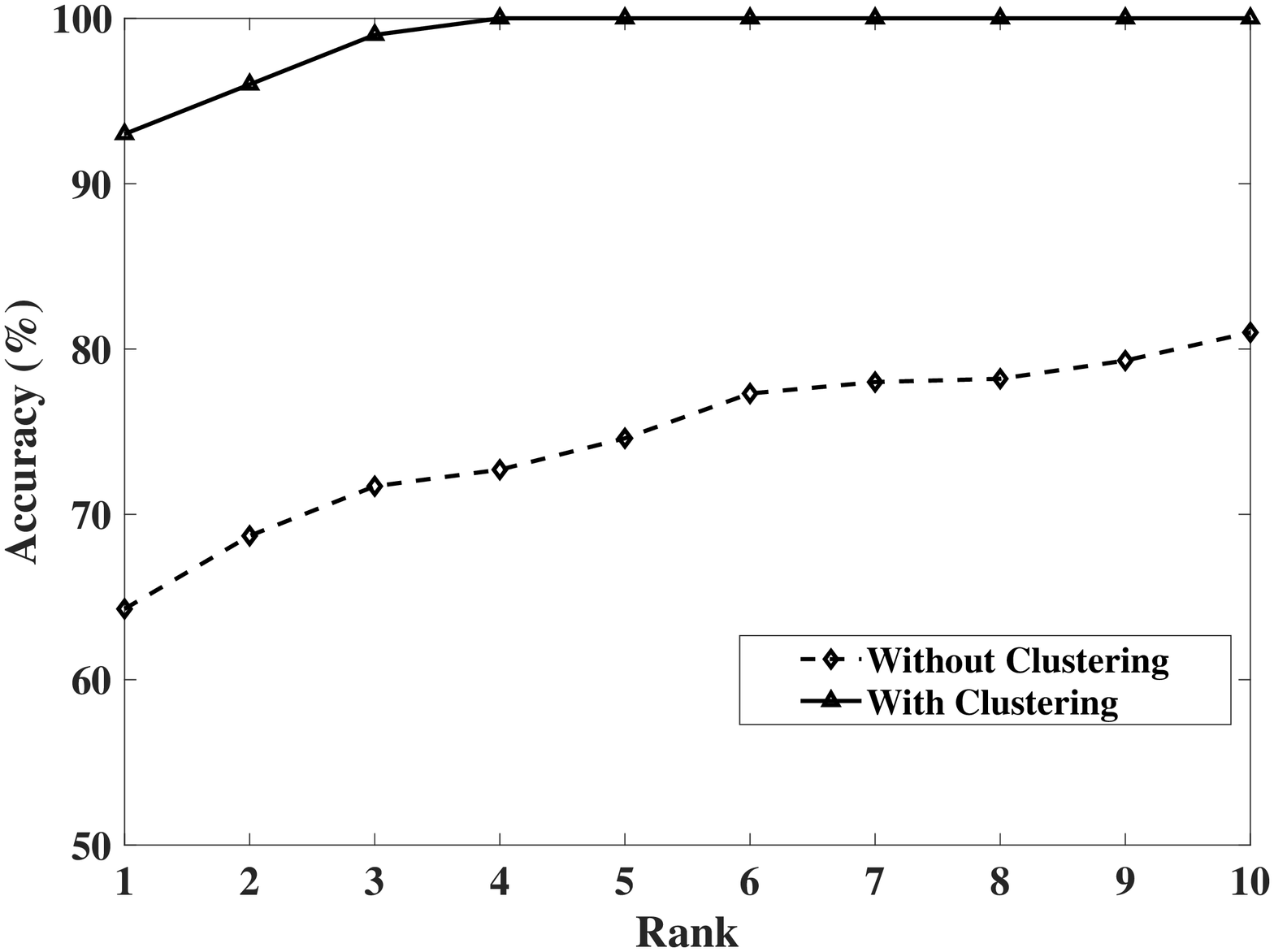}}
	\end{subfigure}
\caption{Cumulative Matching curve showing improvement in accuracy with rank of the proposed algorithm for the two experiments in case of (a) CUHK\_01, (b) CUHK\_03, (c) Market1501 data sets}
\label{cmc_new}
\end{figure*}
From the figures, it can be again observed that the proposed approach always outperforms the state-of-the-art re-identification methods by a large margin. Moreover, our approach achieves the 100\% accuracy mark for all the data sets within a rank of 4. 
As already explained with reference to Table \ref{Table:data1}, the improved performance of the proposed technique is mostly due to the employment of a hierarchical classification process (refer to Section \ref{pa}) in which vastly dissimilar features get eliminated at each level of hierarchy, and only the best matches are retained for further classification at the subsequent level. Another advantage of the use of a hierarchical classification strategy is that the final re-identified result does not depend on the effectiveness of a particular feature. 
Rather, it fuses information from multiple features which enables it to perform better than state-of-the-art approaches those make prediction on the basis on a single feature. 

As explained in Section \ref{tep}, the proposed re-identification algorithm consists of two major components: (i) determining the appropriate cluster/s by utilizing the color information of the test subject and reduce the search space, (ii) comparing the test subject with the elements of the reduced set by means of a Siamese network. In the following experiment, we study the effect of the these individual components on the overall accuracy and response time. Specifically, we report results from two different experiments: (a) rank-based accuracy by including the cluster matching component by setting \emph{K} to 100 and $\mathcal{K}$ to 1, and (b) rank-based accuracy without including the cluster matching component. The same training-test pair as well as the pre-trained Siamese network used in the previous experiments has also been used here. Figures \ref{cmc_new} (i), (ii), (iii) show the CMC curves of rank-wise improvement in re-identification accuracy corresponding to the following three data sets: CUHK\_01, CUHK\_03, Market1501. The average time to re-identify a test subject with and without clustering are 0.232 secs and 0.454 secs, respectively. It is clearly seen from the figure that the clustering step (i.e., the first level of the proposed hierarchical re-identification method) has a profound influence on the final re-identification accuracy. As explained before, the superior performance of the proposed approach with the clustering step is due to the fact that the final prediction using the Siamese network is made by comparing only a small set of closely similar samples and not the entire gallery set.

\section{Conclusions}
In this paper, 
we propose a hierarchical approach for person re-identification in which color histogram based matching is employed at the first level of hierarchy to retain the top few closest matches, and next Siamese network based similarity matching is performed to predict the correct match. Incorporation of the initial color based matching scheme reduces the search space and prevents the Siamese network from getting biased towards a utterly different element of the gallery set. Results on several public data sets demonstrate that the proposed approach outperforms state-of-the-art re-identification techniques by a high margin. Although, as per the scenario of Figure \ref{fig:scenario}, both the walking videos are assumed to be captured from the same view-point, our approach is equally effective for data sets captured from different viewpoints. This is verified from the fact that the proposed approach performs better than each of the existing techniques used in the comparative analysis for each of the six public data sets, most of which consist of multi-view data.  In future, our approach can be extended to improve the accuracy on data set with similar clothing conditions and also perform re-identification in presence of occlusion, or situations when the gallery set evolves continuously.

\section*{Acknowledgments}
The authors would like to acknowledge NVIDIA for supporting their research with the TITAN Xp Graphics processing unit, and all their colleagues who took part in construction of the data set. The authors also express their sincere gratitude to the Head of the Computer Science and Engineering Department and Director of IIT (BHU), Varanasi for their support.


\end{document}